\documentclass{article}

\usepackage[main, final]{neurips_2026}

\usepackage[utf8]{inputenc} 
\usepackage[T1]{fontenc}    
\usepackage{hyperref}       
\usepackage{url}            
\usepackage{booktabs}       
\usepackage{amsfonts}       
\usepackage{nicefrac}       
\usepackage{microtype}      
\usepackage{xcolor}         

\usepackage{amsmath}
\usepackage{bm}
\usepackage{amssymb}
\usepackage{algorithm}
\usepackage{algpseudocode}

\usepackage{amsthm} 
\usepackage{graphicx}
\newtheorem{theorem}{Theorem} 

\usepackage{pifont}
\newcommand{\cmark}{\ding{51}}
\newcommand{\xmark}{\ding{55}}

\usepackage{booktabs}
\usepackage{makecell}
\usepackage{multirow}

\title{Multiscale Supervised Unbalanced Optimal Transport Flow Matching}

\author{%
  Qiangwei Peng\thanks{Equal contribution.} \\
  LMAM and School of Mathematical Sciences \\
  Peking University, Beijing, China \\
  \texttt{qiangweipeng123@gmail.com} \\
  \And
  Lezhi Chen\footnotemark[1] \\
  College of Electrical Engineering \\
  Sichuan University, Sichuan, China \\
  \texttt{chenlezhi\_scu@outlook.com} \\
  \AND
  Peijie Zhou\thanks{Corresponding author. Additional affiliations:
  Center for Quantitative Biology, Peking University, Beijing, China;
  National Engineering Laboratory for Big Data Analysis and Applications,
  Beijing, China; AI for Science Institute, Beijing, China.} \\
  Center for Machine Learning Research \\
  Peking University, Beijing, China \\
  \texttt{pjzhou@pku.edu.cn} \\
}

\begin{document}

\maketitle

\begin{abstract}
  Unbalanced optimal transport (UOT) provides a principled framework for modeling single-cell transitions and birth-death dynamics, but its high computational cost limits scalability to large-scale datasets. Although single-cell data often contain hierarchical annotations and known transition priors, existing UOT approximations rarely exploit this multiscale structure or prior knowledge. We introduce Multiscale Supervised Unbalanced Optimal Transport Flow Matching (MUST-FM), a simulation-free framework that scales UOT by leveraging hierarchical data structure. MUST-FM further supports an optional supervised formulation that incorporates transition priors, such as cell lineages, to guide the learning of displacement fields and mass variations. Experiments show that MUST-FM reduces computational overhead while achieving robust and biologically meaningful trajectory inference, enabling dynamic modeling of atlas-scale single-cell datasets. The code is available at \url{https://github.com/QiangweiPeng/MUST-FM}.
\end{abstract}

\section{Introduction}

Inferring cellular developmental trajectories from unpaired temporal snapshots remains a fundamental challenge in single-cell transcriptomics \citep{chen2022spatiotemporal,waddingot,zheng2017massively}. Optimal transport (OT) provides a principled framework for aligning time-series scRNA-seq distributions \citep{bunne2024optimal, zhang2025review,zhang2025deciphering}: static OT aligns cross-sectional distributions \citep{waddingot, moscot}, whereas dynamic OT reconstructs continuous vector fields \citep{trajectorynet, mioflow, scnode}. However, neural ODE-based dynamic OT methods \citep{chen2018neural} often suffer from high computational costs and training instability due to repeated simulations \citep{yan2019robustness}. Flow matching (FM) \citep{cfm_lipman, albergo2022building}, when coupled with OT, provides a simulation-free solution to dynamic OT problems, thereby establishing a new standard for reconstructing single-cell developmental trajectories \citep{cfm_tong, klein2024genot}.

However, biological systems are intrinsically non-mass-conserving, as proliferation and apoptosis lead to unbalanced distributions over time \citep{TIGON}. The Wasserstein-Fisher-Rao (WFR) metric \citep{liero2016optimal, chizat2018interpolating, kondratyev2016} extends dynamic OT by coupling spatial displacement with mass variation, offering a rigorous geometry for unbalanced dynamics. Several dynamic UOT solvers have been developed using neural ODEs \citep{TIGON, peng2026stvcr, DeepRUOT, sun2025variational}, and recent unbalanced FM methods further avoid ODE integration by jointly regressing transport velocity and growth rate functions \citep{eyring2024unbalancedness, wang2025joint, peng2026wfrfm, wang2026wfr}.

Despite these advances, computing the underlying OT plan remains a major bottleneck when scaling to large datasets. Existing approximation strategies, including Sinkhorn regularization \citep{cuturi2013sinkhorn}, multiscale solvers for balanced OT\citep{gerber2017multiscale, merigot2011multiscale}, low-rank methods \citep{scetbon2022low,scetbon2023unbalanced}, and mini-batch OT \citep{sommerfeld2019optimal,peng2026wfrfm}, improve scalability but largely treat cells as unstructured point clouds. This overlooks a key property of single-cell atlases: as datasets scale to millions of cells, they often come with multi-level annotations and biological transition priors, such as lineage relationships or cell-type compatibility constraints \citep{cao2019single, farrell2018single, packer2019lineage, tabula2020single, la2021molecular}. Although supervised OT can incorporate prior transport information in balanced settings \citep{cang2022supervised}, imposing rigid transition constraints together with strict marginal preservation may lead to infeasible optimization problems.

To address these limitations, we introduce Multiscale Supervised Unbalanced Optimal Transport Flow Matching (MUST-FM). MUST-FM combines multiscale UOT with simulation-free flow matching to exploit hierarchical data structures for scalable dynamic UOT, and further supports an optional supervised formulation that guides transport with multi-level transition priors while avoiding the infeasibility of strict balanced constraints. Our main contributions are:

\begin{itemize}
    \item We propose MUST-FM, a scalable simulation-free framework for dynamic UOT that exploits hierarchical multiscale structure to reduce the cost of global UOT solvers and avoid the structural degradation caused by random mini-batching.
    \item We formulate an optional supervised UOT mechanism that incorporates transition priors across hierarchical scales, enabling feasible prior-guided learning of both displacement and mass variation.
	\item We design a specialized constant-time coupling strategy whose relative cost decreases with data scale and becomes asymptotically negligible during training. This enables MUST-FM to efficiently scale to massive datasets and outperform state-of-the-art baselines in efficiency, robustness, and accuracy for prior-guided trajectory inference.
\end{itemize}

\section{Related Works}

\textbf{Scalable Optimal Transport and Supervised Methods.}
Computing optimal transport on large-scale datasets is computationally prohibitive. To improve scalability, various approximation strategies have been developed, including Sinkhorn regularization \citep{cuturi2013sinkhorn}, multiscale solvers \citep{merigot2011multiscale, gerber2017multiscale}, low-rank approximations \citep{scetbon2022low}, and mini-batch OT \citep{sommerfeld2019optimal,cfm_tong}. Some of these ideas have also been extended to unbalanced settings, such as low-rank UOT \citep{scetbon2023unbalanced} and mini-batch couplings \citep{peng2026wfrfm}. However, these methods generally treat samples as unstructured point clouds and do not explicitly exploit the hierarchical organization often available in biological data. In parallel, supervised OT incorporates known point-to-point transport information to guide coupling construction \citep{cang2022supervised}, but its balanced formulation with strict marginal constraints can become infeasible under rigid transition priors. MUST-FM addresses this gap through a \textbf{scalable multiscale approximation and a supervised formulation designed for unbalanced optimal transport}.

\textbf{Unbalanced Optimal Transport.}
Unbalanced optimal transport relaxes the strict marginal constraints of classical OT \citep{kantorovich1942translocation, benamou2000computational} using divergence penalties \citep{benamou2003numerical, figalli2010optimal, caffarelli2010free} or source terms, making it suitable for systems with varying total mass. A prominent dynamic formulation is the Wasserstein-Fisher-Rao (WFR) geometry, which couples displacement with growth dynamics \citep{liero2016optimal, chizat2018interpolating, kondratyev2016}. For dynamic modeling, neural ODE-based UOT solvers \citep{TIGON, DeepRUOT, peng2026stvcr, sun2025variational} can capture transport and mass variation but suffer from high integration costs. Recent simulation-free flow matching methods, such as VGFM \citep{wang2025joint} and WFR-FM \citep{peng2026wfrfm}, avoid repeated ODE simulation by learning transport and growth fields directly. However, these methods typically operate on flat data representations and do not incorporate multiscale annotations or transition priors. MUST-FM extends this line of work by introducing a \textbf{multiscale and prior-guided} flow matching framework for unbalanced dynamics.

\textbf{Single-cell Trajectory Inference.}
Trajectory inference from time-series scRNA-seq data has motivated many OT-based algorithms, ranging from static alignment methods \citep{waddingot, moscot, halmos2025dest} to continuous dynamic reconstructions using neural ODEs \citep{trajectorynet, scnode, mioflow} and flow matching \citep{cfm_tong, rohbeck2025modeling, klein2024genot}. To capture cell proliferation and apoptosis, recent works increasingly adopt unbalanced formulations \citep{TIGON, DeepRUOT, peng2026stvcr, action_matching, wang2025joint, peng2026wfrfm}. Meanwhile, modern single-cell atlases scale to millions of cells and often contain hierarchical cell-type annotations and biological lineage priors \citep{cao2019single,packer2019lineage}. Current trajectory inference methods, however, rarely use these annotations and priors to guide scalable unbalanced dynamics. MUST-FM complements existing approaches by \textbf{explicitly leveraging multiscale cell-type annotations and transition priors for biologically guided trajectory reconstruction}.

\section{Preliminaries}
We first review the dynamic WFR formulation and the closed-form Dirac-to-Dirac geodesic, which serve as the basis for the multiscale supervised WFR problem and unbalanced flow matching constructions below.

\textbf{Dynamical WFR Problem.}
The WFR metric \citep{chizat2018interpolating, chizat2018unbalanced, liero2018optimal} extends dynamic optimal transport to mass-varying systems by jointly modeling spatial transport and growth. Given two non-negative measures $\mu_0$ and $\mu_1$, the dynamic WFR problem is defined as
\begin{equation}
\label{eq:standard_dynamical_wfr}
\begin{aligned}
\mathrm{WFR}_{\delta}^2(\mu_0,\mu_1)
:= \inf_{\rho,g,\bm{u}}
&\int_0^1\int_{\mathcal{X}}
\frac{1}{2}\Big(\|\bm{u}(\bm{x},t)\|_2^2+\delta^2 |g(\bm{x},t)|^2\Big)
\rho_t(\bm{x})\,\mathrm{d}\bm{x}\mathrm{d}t  \\
\mathrm{s.t.}\quad
&\partial_t\rho+\nabla_{\bm{x}}\cdot(\rho \bm{u})=\rho g,\quad
\rho_0=\mu_0,\ \rho_1=\mu_1 .
\end{aligned}
\end{equation}
Here $\bm{u}$ denotes the transport velocity, $g$ denotes the growth rate, and $\delta$ balances the cost of transport and mass creation or annihilation.

\textbf{Dirac-to-Dirac WFR Geodesic.}
A useful special case of the WFR geometry is the Dirac-to-Dirac problem, whose geodesic and distance admit closed-form expressions \citep{chizat2018interpolating}. For two mass points $m_0\delta_{\bm{x}_0}$ and $m_1\delta_{\bm{x}_1}$, the WFR distance is
\begin{equation}
	\label{eq:dirac_wfr}
	\mathrm{WFR\text{-}DD}_{\delta}^2(m_0\delta_{\bm{x}_0},m_1\delta_{\bm{x}_1})
	=2\delta^2\left(m_0+m_1-2\sqrt{m_0m_1}\,
	\overline{\cos}\Big(\frac{\|\bm{x}_0-\bm{x}_1\|_2}{2\delta}\Big)\right),
\end{equation}
where $\overline{\cos}(x)=\cos(\min\{x,\pi/2\})$. When $\|\bm{x}_0-\bm{x}_1\|_2<\pi\delta$, the corresponding optimal path is the traveling Dirac:
\begin{equation}
	\label{eq:travelling_dirac}
	m(t)=At^2-2Bt+m_0,\quad \bm{u}(t)m(t)=\bm{\omega}_0,
\end{equation}
where
\begin{equation}
	\label{eq:ABw}
	\left\{
	\begin{aligned}
		A&=m_0+m_1-2\sqrt{\frac{m_0m_1}{1+\tau^2}},
		\quad
		B=m_0-\sqrt{\frac{m_0m_1}{1+\tau^2}},\\
		\bm{\omega}_0&=2\delta\tau\sqrt{\frac{m_0m_1}{1+\tau^2}}\,\bm{l},
		\quad
		\tau=\tan\Big(\frac{\|\bm{x}_1-\bm{x}_0\|_2}{2\delta}\Big).
	\end{aligned}
	\right.
\end{equation}
Here $\bm{l}$ is the unit vector pointing from $\bm{x}_0$ to $\bm{x}_1$.

\section{Supervised Unbalanced Optimal Transport}
\label{sec:supervised_uot}

We now introduce the supervised extension of WFR by incorporating biological transition priors into the admissible transport paths. The goal is to preserve the unbalanced transport geometry of WFR while restricting mass movement to biologically plausible endpoint transitions.

\textbf{Supervised Dynamical WFR.}
Let $M$ be a binary transition mask, where $M(\bm{x},\bm{y})=1$ indicates that the transition from $\bm{x}$ to $\bm{y}$ is biologically admissible. We restrict the endpoint pairs of WFR trajectories by defining the admissible path space $\Omega_M=\{\omega\in\mathcal{C}([0,1],\mathcal{X})\mid M(\omega(0),\omega(1))=1\}.$
Using the superposition principle, we represent $\rho_t$ through a non-negative Lagrangian path measure $\boldsymbol{\eta}$. The supervised dynamical WFR problem is then defined as
\begin{equation}
\label{eq:supervised_dynamical_wfr}
\begin{aligned}
\text{S-WFR}_{\delta}^2&(\mu_0,\mu_1; M)
:= \inf_{\rho,g,\bm{u},\boldsymbol{\eta}}
\int_0^1\int_{\mathcal{X}}
\frac{1}{2}\Big(\|\bm{u}(\bm{x},t)\|_2^2+\delta^2 |g(\bm{x},t)|^2\Big)
\rho_t(\bm{x})\,\mathrm{d}\bm{x}\mathrm{d}t  \\
\mathrm{s.t.}\quad
&\partial_t\rho+\nabla_{\bm{x}}\cdot(\rho \bm{u})=\rho g,\quad
\rho_0=\mu_0,\ \rho_1=\mu_1, \\
&\rho_t=(e_t)_{\sharp}(w_t\boldsymbol{\eta}),\quad
w_t(\omega)=\exp\left(\int_0^t g(\omega(s),s)\,\mathrm{d}s\right),\quad
\mathrm{supp}(\boldsymbol{\eta})\subseteq\Omega_M .
\end{aligned}
\end{equation}
Here $(e_t)_{\sharp}$ denotes the push-forward of the evaluation map at time $t$. Compared with the standard WFR problem in Eq.~\ref{eq:standard_dynamical_wfr}, the additional support constraint on $\boldsymbol{\eta}$ enforces that mass can only move along biologically admissible endpoint transitions specified by $M$.

\textbf{Supervised Static Semi-coupling.}
To construct the coupling for flow matching, we convert the supervised dynamic constraint into a static Kantorovich formulation. The Benamou-Brenier equivalence for unsupervised WFR \citep{chizat2018unbalanced, liero2018optimal} naturally extends to this setting: restricting the path measure $\boldsymbol{\eta}$ to $\Omega_M$ is equivalent to restricting the static semi-coupling to valid endpoint pairs. The static supervised WFR problem is
\begin{equation}
	\label{eq:static_wfr}
	\text{S-WFR}_{\delta}^2(\mu_0,\mu_1; M) = \inf_{(\gamma_0,\gamma_1)\in \Gamma_M(\mu_0, \mu_1)} \int_{\mathcal{X}^2}  \, \text{WFR-DD}_{\delta}^2(\gamma_0(\bm{x},\bm{y})\delta_{\bm{x}},\gamma_1(\bm{x},\bm{y})\delta_{\bm{y}})\mathrm{d} \bm{x}\mathrm{d} \bm{y},
\end{equation}
where the feasible set $\Gamma_M$ is
\begin{equation}
\label{eq:feasible_set}
\begin{aligned}
\Gamma_M(\mu_0,\mu_1)
:=\big\{(\gamma_0,\gamma_1)\in\mathcal{M}_+(\mathcal{X}^2)^2:
&\int_{\mathcal{X}}\gamma_0(\bm{x},\bm{y})\,d\bm{y}=\mu_0(\bm{x}),\
\int_{\mathcal{X}}\gamma_1(\bm{x},\bm{y})\,d\bm{x}=\mu_1(\bm{y}),\\
&\mathrm{supp}(\gamma_0),\mathrm{supp}(\gamma_1)\subseteq S_M
\big\},
\end{aligned}
\end{equation}
with $S_M=\{(\bm{x},\bm{y}):M(\bm{x},\bm{y})=1\}$.
For each pair $(\bm{x},\bm{y})$, $\gamma_0(\bm{x},\bm{y})$ and $\gamma_1(\bm{x},\bm{y})$ denote the initial mass sent from $\bm{x}$ and the final mass received by $\bm{y}$, respectively. In the unbalanced setting, these two masses may differ, while the support constraint forbids transport along biologically invalid transitions.

\textbf{Equivalent Supervised Optimal Entropy-Transport.}
To efficiently solve the supervised WFR coupling, we recast the static problem as a masked optimal entropy-transport (OET) problem, where the prior mask $M$ restricts the support of the coupling:
\begin{equation}
	\label{eq:OET}
	\begin{aligned}
		\text{S-WFR}_{\delta}^2(\mu_0,\mu_1; M) &= 2\delta^2\inf_{\gamma\in\mathcal{M}_M} \Bigg\{\int_{\mathcal{X}^2}  \, -2\operatorname{ln}\operatorname{\overline{\cos}}\Big(\frac{\Vert \bm{x}-\bm{y}\Vert_2}{2\delta}\Big)\gamma(\bm{x},\bm{y})\mathrm{d} \bm{x}\mathrm{d} \bm{y}\\&+\operatorname{KL}\Big(\int_\mathcal{X}\gamma(\bm{x},\bm{y})\mathrm{d} \bm{y}\Vert \mu_0(\bm{x})\Big)+\operatorname{KL}\Big(\int_\mathcal{X}\gamma(\bm{x},\bm{y})\mathrm{d} \bm{x}\Vert \mu_1(\bm{y})\Big)\Bigg\}
	\end{aligned}
\end{equation}
where $\mathcal{M}_M=\{\gamma\in\mathcal{M}_+(\mathcal{X}^2):\mathrm{supp}(\gamma)\subseteq S_M\}$.
The relation between the OET coupling $\gamma$ and the WFR semi-coupling $(\gamma_0,\gamma_1)$ is well established in the unsupervised setting \citep{liero2018optimal,peng2026wfrfm}. In our supervised setting, the mask only restricts the admissible support, so the optimal semi-coupling admits the same normalization form.
\begin{theorem}
	\label{thm:semi-coupling}
	Let $\gamma$ solve the supervised OET problem in Eq.~\ref{eq:OET}. Then the semi-coupling defined by
	$\gamma_0(\bm{x},\bm{y})=\frac{\gamma(\bm{x},\bm{y})}{\int_\mathcal{X}\gamma(\bm{x},\bm{z})\,d\bm{z}}\mu_0(\bm{x})$
	and
	$\gamma_1(\bm{x},\bm{y})=\frac{\gamma(\bm{x},\bm{y})}{\int_\mathcal{X}\gamma(\bm{z},\bm{y})\,d\bm{z}}\mu_1(\bm{y})$
	solves the static supervised WFR problem in Eq.~\ref{eq:static_wfr}.
\end{theorem}

\section{Multiscale Unbalanced Optimal Transport via Hierarchical Supervision}\label{sec:MS-UOT}

While the previous section establishes a theoretical foundation for supervised UOT in continuous spaces, directly solving the Kantorovich or OET problem on single-cell datasets with millions of observations is computationally prohibitive \citep{moscot}. Modern single-cell atlases, however, are rarely unstructured point clouds: they are often organized by hierarchical annotations, from coarse lineages to fine-grained cell subtypes, and are accompanied by transition priors at multiple resolutions, such as lineage relationships, cell-type compatibility, or subtype-level constraints \citep{wang2025drosophila,chen2022spatiotemporal,cao2019single,qiu2024spatiotemporal,kern2025merfish+}. We exploit this multiscale structure by projecting continuous measures onto discrete hierarchical representations and solving supervised UOT in a coarse-to-fine manner, thereby reducing computational complexity while preserving biologically admissible transitions across scales.

\textbf{Multiscale Discrete Representation.}
For source and target single-cell data, we assume a hierarchy with $L$ annotation levels, from the coarsest ($l=1$) to the finest ($l=L$). At level $l$, we approximate the corresponding empirical measures by weighted centroid measures:
\begin{equation}
	\label{eq:discrete_measure}
	\hat{\mu}_0^{(l)} = \sum_{i=1}^{K_0^{(l)}} w_{0,i}^{(l)} \delta_{\bm{c}_{0,i}^{(l)}}, \quad 
	\hat{\mu}_1^{(l)} = \sum_{j=1}^{K_1^{(l)}} w_{1,j}^{(l)} \delta_{\bm{c}_{1,j}^{(l)}} .
\end{equation}
Here $K_0^{(l)},K_1^{(l)}$ are the numbers of clusters at level $l$, and $\bm{c}_{0,i}^{(l)}, \bm{c}_{1,j}^{(l)} \in \mathcal{X}$ are the corresponding centroids. To preserve mass variation in WFR dynamics, the weights $w_{0,i}^{(l)}$ and $w_{1,j}^{(l)}$ are absolute, unnormalized cell counts rather than probabilities.

\textbf{Hierarchical Mask Construction.}
At each level $l$, the discrete transport plan is constrained by a binary mask $M^{(l)}\in\{0,1\}^{K_0^{(l)}\times K_1^{(l)}}$. To fully leverage both domain knowledge and data-driven topology, we construct $M^{(l)}$ by intersecting two distinct sets of constraints:

\textbf{1. Biological Knowledge Prior ($B^{(l)}$):} Biological annotations often come with prior knowledge about admissible transitions, including lineage relationships, or compatibility between cell types. We encode such information by a binary matrix $B^{(l)}$, where $B_{ij}^{(l)}=0$ rules out a biologically forbidden transition from source cluster $i$ to target cluster $j$, and $B_{ij}^{(l)}=1$ leaves the transition admissible.

\textbf{2. Coarse-to-Fine Guidance ($H^{(l)}$):} For $l>1$, we use the OET coupling $\gamma^{(l-1)}$ from the coarser level to prune unlikely fine-level transitions. Let $\mathrm{pa}(i)$ denote the parent of cluster $i$ at level $l-1$. Given a threshold $\epsilon\in(0,1)$, we allow transitions between fine clusters only if their parent-level transition probability exceeds $\epsilon$. The hierarchical mask is defined as $H_{ij}^{(l)}=\mathbb{I}\!\left(\gamma_{\mathrm{pa}(i),\mathrm{pa}(j)}^{(l-1)}/w_{0,\mathrm{pa}(i)}^{(l-1)}\ge\epsilon\right).$

The final effective mask for level $l$ is the element-wise logical AND of both constraints:
\begin{equation}
	\label{eq:final_mask}
	M^{(l)} = B^{(l)} \odot H^{(l)}.
\end{equation}
For the coarsest level $l=1$, the hierarchical mask is inactive, yielding $M^{(1)} = B^{(1)}$. 

\textbf{Multiscale Supervised OET Objective.}
Given the discrete representations $\hat{\mu}_0^{(l)}$ and $\hat{\mu}_1^{(l)}$ with hierarchical mask $M^{(l)}$, we discretize the supervised OET formulation at each level $l$. Let $\mathbf{C}^{(l)}\in\mathbb{R}^{K_0^{(l)}\times K_1^{(l)}}$ be the WFR cost matrix, with entries
$C_{ij}^{(l)}=-2\ln\overline{\cos}\!\left(\frac{\|\bm{c}_{0,i}^{(l)}-\bm{c}_{1,j}^{(l)}\|_2}{2\delta}\right)$.
The multiscale supervised OET objective is
\begin{equation}
\label{eq:discrete_OET}
\min_{\gamma^{(l)}\in\mathbb{R}_+^{K_0^{(l)}\times K_1^{(l)}},\ \gamma_{ij}^{(l)}=0\ \mathrm{if}\ M_{ij}^{(l)}=0}
\sum_{i=1}^{K_0^{(l)}} \sum_{j=1}^{K_1^{(l)}} C_{ij}^{(l)} \gamma_{ij}^{(l)}
+\operatorname{KL}\!\left(\gamma^{(l)}\mathbf{1}\Vert w_0^{(l)}\right)
+\operatorname{KL}\!\left((\gamma^{(l)})^\top\mathbf{1}\Vert w_1^{(l)}\right).
\end{equation}
Solving Eq.~\ref{eq:discrete_OET} from $l=1$ to $L$ yields increasingly fine-grained, biologically constrained OET couplings. As the mask is propagated downward, the effective transition matrix becomes increasingly sparse at fine resolutions. We exploit this sparsity by modifying the OET solver in the Python Optimal Transport (POT) library \citep{flamary2021pot} to skip computations on masked elements, reducing the cost of coupling construction on large single-cell datasets. The finest-level coupling $\gamma^{(L)}$ is then used as endpoint supervision for the continuous flow matching model.

\section{Simulation-Free Training for MUST-FM}

To reconstruct continuous dynamics from discrete single-cell snapshots, we learn a time-dependent velocity field $\bm{v}_{\bm{\theta}}(\bm{x},t)$ and a growth rate function $g_{\bm{\phi}}(\bm{x},t)$. Because direct optimization over marginal unbalanced dynamics is intractable, we build upon recent advances in unbalanced flow matching \citep{peng2026wfrfm}, which extends simulation-free conditional flow matching (CFM) to mass-varying trajectories.

\textbf{Tractable Unbalanced Flow Matching with Traveling Gaussian.}
We bypass intractable marginal dynamics by defining conditional paths over endpoint pairs sampled from the semi-coupling $\gamma_0(\bm{x}_0,\bm{x}_1)$. Each Conditional Gaussian Measure Path (CGMP) is factorized into a mass term and a probability density:
\begin{equation}
	\label{eq:CGMP}
	\rho_t(\bm{x} \vert \bm{x}_0, \bm{x}_1) =
	m_t(\bm{x}_0, \bm{x}_1)
	\mathcal{N}\big(\bm{x}\mid \bm{\eta}_t(\bm{x}_0,\bm{x}_1), \sigma_t^2\mathbf{I}\big),
\end{equation}
which induces the conditional vector field
$\bm{u}_t=\frac{\sigma_t'}{\sigma_t}(\bm{x}-\bm{\eta}_t)+\bm{\eta}_t'$
and growth rate $g_t=\partial_t\ln m_t$.

To make the conditional path consistent with WFR geometry, we align it with the analytical WFR geodesic between two Dirac masses \citep{chizat2018interpolating}. Following recent flow matching formulations \citep{cfm_tong,peng2026wfrfm}, we use a \textit{Traveling Gaussian} path whose mean trajectory $\bm{\eta}_t$ and mass evolution $m_t$ are given by:
\begin{equation}
	\label{eq:travelling_gaussian_mean}
	\begin{aligned}
		\bm{\eta}_t(\bm{x}_0, \bm{x}_1) &= \bm{x}_0 + \frac{\bm{\omega}_0}{\sqrt{m_0 A - B^2}} \left( \arctan \frac{At - B}{\sqrt{m_0 A - B^2}} - \arctan \frac{-B}{\sqrt{m_0 A - B^2}} \right), \\
		m_t(\bm{x}_0, \bm{x}_1) &= At^2 - 2Bt + m_0,
	\end{aligned}
\end{equation}
where $A$, $B$, and $\bm{\omega}_0$ are geometric constants defined in Eq.~\ref{eq:ABw}. We train the neural networks with the mass-weighted Conditional Unbalanced Flow Matching (CUFM) objective:
\begin{equation}
	\label{eq:CUFM_simplified}
	\mathcal{L}_{\mathrm{CUFM}}
	=
	\mathbb{E}_{t\sim\mathcal{U}[0,1],\,(\bm{x}_0,\bm{x}_1)\sim\gamma_0,\,\bm{x}\sim\mathcal{N}(\bm{\eta}_t,\sigma_t^2\mathbf{I})}
	\Big[
	\big(
	\|\bm{v}_{\bm{\theta}}-\bm{u}_t\|_2^2
	+\kappa (g_{\bm{\phi}}-g_t)^2
	\big)
	m_t(\bm{x}_0,\bm{x}_1)
	\Big].
\end{equation}
Minimizing this conditional loss guarantees the recovery of the target marginal dynamics\citep{peng2026wfrfm}.

\textbf{Multiscale Supervision via Hierarchical Semi-coupling.}
The generative fidelity of the CUFM framework fundamentally depends on the choice of the coupling plan $\gamma_0$ and the corresponding mass boundary conditions $m_0, m_1$. Instead of relying on unsupervised heuristics, we uniquely drive the generative flow using the finest-level supervised semi-coupling $(\gamma_0^{(L)}, \gamma_1^{(L)})$ derived in Section \ref{sec:MS-UOT}. 

This hierarchical semi-coupling inherently encapsulates both the established biological knowledge priors ($B^{(L)}$) and the coarse-to-fine topological restrictions. For a coupled pair $(\bm{x}_0, \bm{x}_1)$ sampled from this supervised plan, we establish the relative mass boundary conditions for the Traveling Gaussian as $m_0(\bm{x}_0,\bm{x}_1)=1$ and 
$m_1(\bm{x}_0,\bm{x}_1)=\gamma_1^{(L)}(\bm{x}_0,\bm{x}_1)/\gamma_0^{(L)}(\bm{x}_0,\bm{x}_1)$.
By enforcing these boundary conditions, the continuous flow matching model is explicitly instructed to learn cellular proliferation or apoptosis rates that strictly respect the macroscopic mass variations dictated by the multiscale lineage topology.

\textbf{Scalability via Finest-Level Within-Block Independent Coupling Lifting.}
Although hierarchical masking reduces the OET search space, solving an exact OET problem at single-cell resolution can still be costly for atlas-scale datasets. Moreover, biological transition priors are usually specified at interpretable annotation levels, such as cell types, subtypes, or lineage groups, rather than individual cells. We therefore solve the supervised OET problem only up to the penultimate level $L-1$, where transition blocks remain biologically meaningful, and use the finest level $L$ only to lift the learned coarse transport structure. This yields a final-step within-block independent approximation: each penultimate-level transport block is redistributed over its finest-level source and target children according to their empirical weights.

Let $\gamma^{(L-1)}\in\mathbb{R}_+^{K_0^{(L-1)}\times K_1^{(L-1)}}$ be the penultimate-level OET coupling. For each finest-level source element $i$ and target element $j$, let $I=\operatorname{pa}_0(i)$ and $J=\operatorname{pa}_1(j)$ denote their parent clusters at level $L-1$. Let $\mathcal{C}_0(I)$ and $\mathcal{C}_1(J)$ be the corresponding child sets, with parent weights
$w_{0,I}^{(L-1)}=\sum_{i\in\mathcal{C}_0(I)}w_{0,i}^{(L)}$ and
$w_{1,J}^{(L-1)}=\sum_{j\in\mathcal{C}_1(J)}w_{1,j}^{(L)}$.
We lift the coupling by
\begin{equation}
	\label{eq:coarse_to_fine_lifting}
	\tilde{\gamma}_{ij}^{(L)}
	=
	\gamma_{IJ}^{(L-1)}
	\frac{w_{0,i}^{(L)}}{w_{0,I}^{(L-1)}}
	\frac{w_{1,j}^{(L)}}{w_{1,J}^{(L-1)}},
	\quad
	I=\operatorname{pa}_0(i),\ J=\operatorname{pa}_1(j).
\end{equation}
This construction preserves the mass of each coarse block, since
$\sum_{i\in\mathcal{C}_0(I)}\sum_{j\in\mathcal{C}_1(J)}\tilde{\gamma}_{ij}^{(L)}=\gamma_{IJ}^{(L-1)}$.

In practice, when flow matching only requires endpoint samples, we do not need to explicitly store the dense finest-level coupling or semi-coupling matrix. Instead, we store the level-$L-1$ coupling and sample level-$L$ pairs implicitly: first sample a parent block $(I,J)$, then sample child indices $i\in\mathcal{C}_0(I)$ and $j\in\mathcal{C}_1(J)$ according to the normalized empirical weights within the block. As shown in Appendix~\ref{app:implicit_finest_sampling}, this implicit procedure samples from the same lifted finest-level pair distribution, further reducing memory and computation.

\section{MUST-FM Workflow for Trajectory Inference}
\label{sec:MUST-FM Workflow}

\begin{figure}[h]
	\centering
	\includegraphics[width=1.0\linewidth]{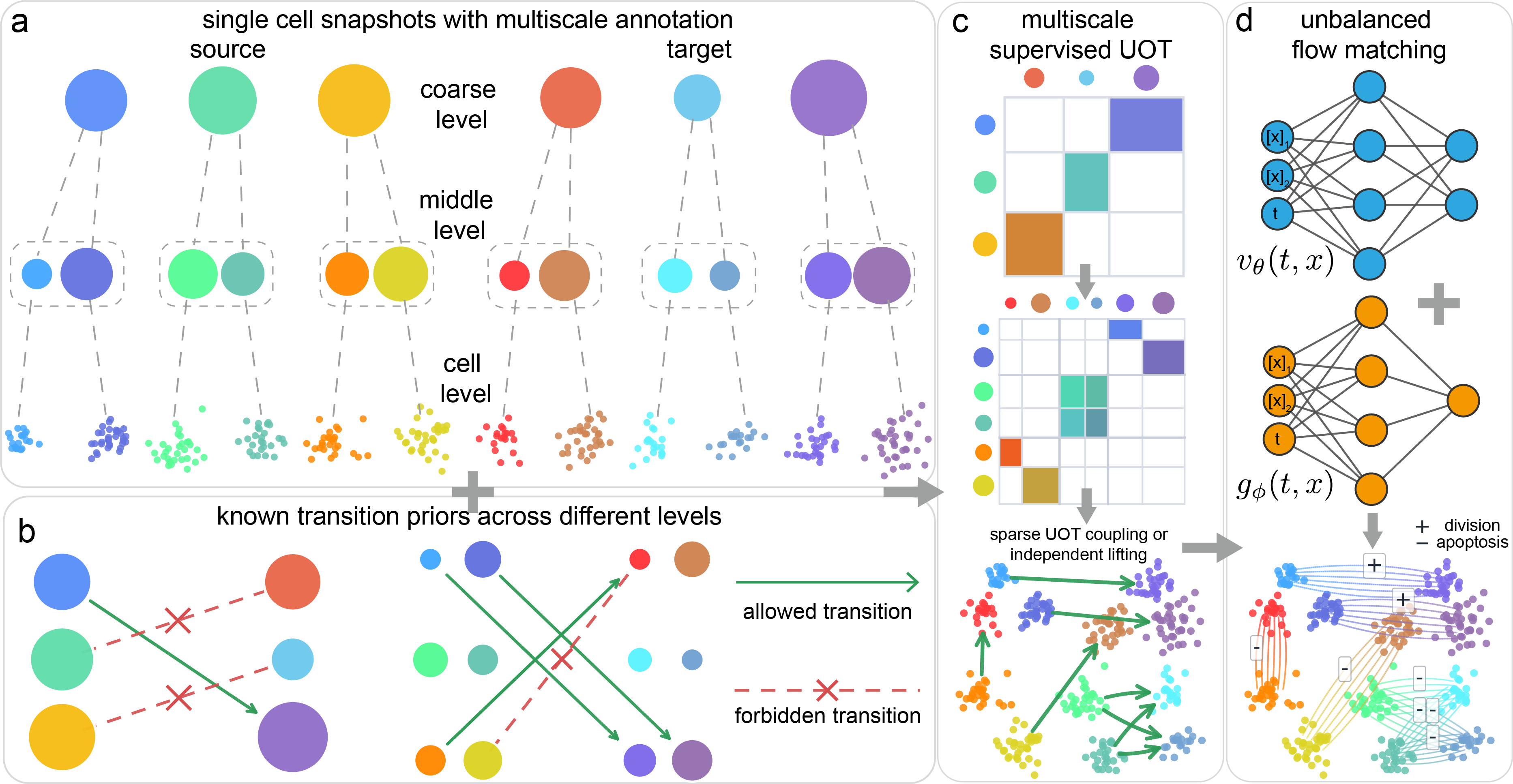}
	\caption{\textbf{MUST-FM workflow.} 
    (a) Multiscale annotations represent source and target single-cell snapshots.
    (b) Biological priors define allowed transitions across scales.
    (c) Multiscale supervised UOT computes sparse prior-guided couplings.
    (d) These couplings supervise simulation-free unbalanced flow matching to learn continuous transport and growth dynamics.}
	\label{fig:overview}
\end{figure}

In this section, we summarize MUST-FM as a practical workflow for reconstructing continuous developmental trajectories from multi-time scRNA-seq data (Figure~\ref{fig:overview}).

\textbf{Multi-time Point WFR Formulation.}
Given samples from distributions $\mu_{t_i}$ at $K+1$ time points $t_0<\cdots<t_K$, we aim to reconstruct the continuous developmental evolution. Following prior work \citep{peng2026wfrfm}, the multi-time WFR objective decomposes into independent pairwise WFR problems, so we construct hierarchical couplings separately for each interval $[t_k,t_{k+1}]$. For dynamics learning, however, we do not train separate models for each interval. Instead, we use time-conditioned neural networks to parameterize a single vector field $\bm{v}_{\bm{\theta}}(\bm{x},t)$ and growth rate $g_{\bm{\phi}}(\bm{x},t)$ over the full temporal domain, enabling parameter sharing, smooth trajectories, and temporal generalization.

\textbf{MUST-FM based on Multiscale Supervision.}
Flow matching requires endpoint pairs sampled from couplings between adjacent time points $(\mu_{t_k},\mu_{t_{k+1}})$. Previous methods \citep{wang2025joint,cfm_tong,peng2026wfrfm} often use random mini-batching to reduce the cost of exact OT. However, random mini-batching can obscure global biological structure and does not explicitly enforce macroscopic topology or known transition priors. We instead use the hierarchical OET solver from Section~\ref{sec:MS-UOT}. For each interval $[t_k,t_{k+1}]$, transition constraints are propagated from coarse annotations to the penultimate level $L-1$, yielding biologically admissible transport blocks. Starting from this penultimate-level solution, we provide two options for handling the finest level $L$:

\textbf{1. Exact Sparse OET at the finest level:} Continue the coarse-to-fine procedure to level $L$ and solve the discrete OET problem only over the finest-level unmasked entries to obtain an optimized finest-level coupling.
    
\textbf{2. Final-Step Within-Block Independent Coupling Lifting:} Stop the OET solve at level $L-1$ and lift each penultimate-level transport block to the finest level by redistributing mass according to empirical child weights (Eq.~\ref{eq:coarse_to_fine_lifting}). When sampling endpoint pairs for flow matching training, the finest-level coupling need not be explicitly stored; see Appendix~\ref{app:implicit_finest_sampling}.

Both strategies preserve the prescribed biological transition topology and provide guidance for the continuous flow model without relying on random mini-batching. The first option further optimizes fine-scale transport geometry, whereas the second trades within-block microscopic optimality for improved scalability. The complete training procedure for multi-time point trajectory inference is provided in Appendix~\ref{subsec:pseudocode} as Algorithm~\ref{alg:MUST-FM}.

\section{Experimental Results}
\label{sec:experimental_results}

We organize our experiments around four questions that evaluate MUST-FM from coupling construction to trajectory learning and scalability. \textbf{Q1} asks whether exploiting intrinsic multiscale structure can improve both the accuracy and efficiency of coupling computation. \textbf{Q2} evaluates whether the continuous flow learned by MUST-FM can successfully transport the initial distribution to subsequent target distributions under the restricted hierarchical state space. \textbf{Q3} examines whether incorporating known transition priors improves interpolation accuracy at unseen time points. \textbf{Q4} assesses the scalability of MUST-FM on large-scale datasets. Additional implementation details, including computational resources and hyperparameter settings, are provided in Appendix~\ref{app:E}.

\textbf{Coupling Accuracy under Multiscale Structure (Q1).} We evaluate coupling accuracy on the multiscale synthetic dataset described in Appendix~\ref{app:balanced_synthetic_data}. This dataset is constructed with an explicit ground-truth coupling, allowing us to directly assess whether each method recovers the known point-, micro-, and macro-level transport structure. We compare exact OT, mini-batch OT \citep{sommerfeld2019optimal}, low-rank OT \citep{scetbon2022low}, and two variants of our methods: a sparse exact finest-level version and a final-step within-block independent lifting version. Exact OT is computed using the POT library \citep{flamary2021pot} with the maximum number of iterations increased to $20\times$ the default value. We report the relative objective-value gap with respect to the ground-truth, point-, micro-, and macro-level accuracy, and computation time. The formulas used to compute these metrics can be found in Appendix~\ref{app:metric}. 

As shown in Table~\ref{tab:q1_multiscale_coupling}, MUST-FM (Sparse) almost perfectly recovers the known point-level correspondence while achieving lower runtime than exact OT. MUST-FM (Independent) does not target exact point-level matching, but preserves the correct micro- and macro-level transitions with much lower computation time. This is important in single-cell dynamics, where cell-type transitions are often more informative than exact point-level matches. To further understand when such multiscale approximations remain reliable, we provide a theoretical analysis and additional controlled experiments in Appendix~\ref{app:D}. Specifically, we derive a sufficient condition for hierarchical pruning to preserve important fine-scale transitions and an approximation error bound for independent lifting in Appendix~\ref{app:D1}. The experiments in Appendix~\ref{app:D2} further show that increasing within-cluster variance or decreasing micro-cluster separation degrades point- and micro-level coupling accuracy, while macro-level transitions remain largely preserved.

\begin{table}[t]
    \centering
    \setlength{\tabcolsep}{2.5pt}
    \caption{\textbf{Coupling accuracy on the synthetic multiscale dataset (Q1).}}
    \label{tab:q1_multiscale_coupling}
    \begin{tabular}{lccccc}
        \toprule
        Method & Obj. Gap (\%) $\downarrow$ & Point Acc. $\uparrow$ & Micro Acc. $\uparrow$ & Macro Acc. $\uparrow$ & Time (s) $\downarrow$ \\
        \midrule
        Ground Truth & 0.00 & 1.000 & 1.000 & 1.000 & -- \\
        \midrule
        Exact OT  &\underline{2.35e-5} & \underline{0.268} & 0.510 & \textbf{1.000} & 89.91 \\
        Mini-batch OT \citep{sommerfeld2019optimal} & 1.50e-1 & 0.002 & 0.328 & 0.988 & \underline{8.86} \\
        Low-rank OT \citep{scetbon2022low} & 1.35e+1 & 0.000 & 0.076 & 0.695 & 151.92 \\
        \midrule
        MUST-FM (Sparse) & \textbf{1.78e-14} & \textbf{0.999} & \textbf{1.000} & \textbf{1.000} & 10.50 \\
        MUST-FM (Independent) & 3.99e-2 & 0.001 & \textbf{1.000} & \textbf{1.000} & \textbf{0.025} \\
        \bottomrule
    \end{tabular}
\end{table}

\begin{table*}[t]
    \centering
    \setlength{\tabcolsep}{4pt}
    \caption{\textbf{Transport accuracy across datasets (Q2).}}
    \label{tab:q2_transport_accuracy}
    \begin{tabular}{lcccccccc}
        \toprule
        \multirow{2}{*}{Method}
        & \multicolumn{2}{c}{Gene 2D}
        & \multicolumn{2}{c}{Gaussian 1000D}
        & \multicolumn{2}{c}{Mouse 2D}
        & \multicolumn{2}{c}{EB 50D} \\
        \cmidrule(lr){2-3}
        \cmidrule(lr){4-5}
        \cmidrule(lr){6-7}
        \cmidrule(lr){8-9}
        & $W_1$ $\downarrow$ & RME $\downarrow$
        & $W_1$ $\downarrow$ & RME $\downarrow$
        & $W_1$ $\downarrow$ & RME $\downarrow$
        & $W_1$ $\downarrow$ & RME $\downarrow$ \\
        \midrule
        VGFM \citep{wang2025joint} & 0.046 & 0.006 & 3.010 & \textbf{0.037} & 0.105 & 0.031 & 8.891 & 0.049 \\
        WFR-FM \citep{peng2026wfrfm} & 0.019 & \textbf{0.001} & \underline{2.233} & \underline{0.044} &
        \textbf{0.042} & \textbf{0.001} & 8.892 & \underline{0.005} \\
        DeepRUOT \citep{DeepRUOT} & 0.043 & 0.017 & 3.785 & 0.303 & 0.139 & 0.171 & 9.092 & 0.312 \\
        TIGON \citep{TIGON} & 0.045 & 0.014 & 2.263 & 0.127 & 0.328 & 0.151 & 9.415 & 0.092 \\
        \midrule
        
        MUST-FM (Sparse) & \textbf{0.016} & \textbf{0.001} & \textbf{2.212} & 0.053 & \textbf{0.042}  & \textbf{0.001} & \textbf{6.794} & \textbf{0.003} \\
        MUST-FM (Independent) & \textbf{0.016} & \textbf{0.001} & 3.035 & 0.194 &  0.051 & \textbf{0.001} & \underline{6.817} & 0.010 \\
        \bottomrule
    \end{tabular}
\end{table*}

\textbf{Transport Accuracy under Hierarchical Constraints (Q2).}
We next evaluate whether the continuous flow learned by MUST-FM can transport the initial distribution to subsequent target distributions despite the restricted hierarchical state space. We compare MUST-FM with representative unbalanced trajectory and flow matching methods, including VGFM \citep{wang2025joint}, WFR-FM \citep{peng2026wfrfm}, DeepRUOT \citep{DeepRUOT}, and TIGON \citep{TIGON}. The evaluation includes Gene 2D and Gaussian 1000D, two widely used synthetic benchmarks adopted in previous unbalanced trajectory modeling studies \citep{DeepRUOT,peng2026wfrfm,wang2025joint}, as well as two real biological datasets: mouse hematopoiesis data (Mouse) \citep{weinreb2020lineage} and human embryoid bodies (EB) \citep{moon2019visualizing} datasets. We report the 1-Wasserstein distance ($W_1$) to measure transport accuracy and relative mass error (RME) to evaluate mass prediction. 

As shown in Table~\ref{tab:q2_transport_accuracy}, MUST-FM achieves competitive or superior performance across both synthetic and real biological datasets while enforcing hierarchical transition constraints. In particular, MUST-FM (Sparse) attains strong transport accuracy and mass prediction, and the final-step independent lifting variant often retains comparable performance despite avoiding an exact finest-level coupling solve. These results indicate that hierarchical supervision does not hinder continuous transport learning, and that independent lifting provides a scalable approximation with little loss in downstream flow accuracy.

\begin{table}[t]
    \centering
    \caption{\textbf{Leave-one-out interpolation on MOCA dataset \citep{cao2019single} (Q3).} We report $W_1$ ($\downarrow$) at each held-out time point.}
    \label{tab:q3_moca_prior}
    \begin{tabular}{lcccc}
        \toprule
        Method & E10.5 & E11.5 & E12.5 & Average \\
        \midrule
        I-CFM \citep{cfm_lipman}  & 1.698 & 1.633 &	1.816 &	1.716\\
        OT-CFM \citep{cfm_tong} & 1.360 &	1.430 &	1.340 &	1.377\\
        \midrule
        MUST-FM w/o Priors & 1.370 & 1.429 & \textbf{1.252} & 1.350\\
        MUST-FM w/ Priors  & \textbf{1.326} & \textbf{1.364} & 1.318 & \textbf{1.336}\\
        \bottomrule
    \end{tabular}
\end{table}

\begin{figure}[t]
    \centering
    \includegraphics[width=1.0\linewidth]{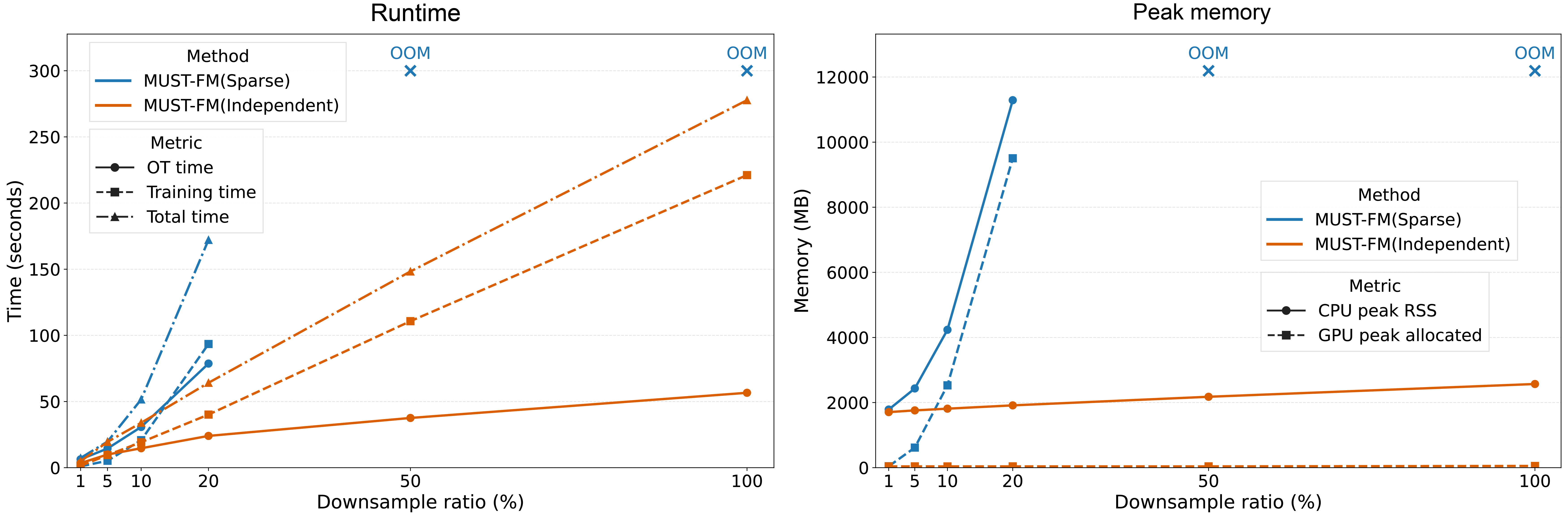}
    \caption{\textbf{Scalability on MOCA \citep{cao2019single} dataset (Q4).}
    Runtime and memory usage of MUST-FM under different subsampling ratios.}
    \label{fig:q4_scalability}
\end{figure}

\textbf{Prior-Guided Interpolation on Large-scale Atlas Datasets (Q3).}
We next evaluate whether biological transition priors improve interpolation at unseen developmental stages on the Mouse Organogenesis Cell Atlas (MOCA) \citep{cao2019single}. MOCA contains approximately $1.3$ million cells across five developmental stages, E9.5, E10.5, E11.5, E12.5, and E13.5, with two-level annotations consisting of $38$ major cell types and $655$ minor cell types. Based on the developmental trajectories reported in the original study, we construct admissible transition priors at both annotation levels.


Due to the atlas scale of MOCA, computing or storing full cell-level couplings, as required by existing proliferation-aware flow matching trajectory methods, is computationally impractical. We therefore evaluate the final-step independent lifting variant of MUST-FM, which avoids explicitly solving the finest-level coupling, alongside two scalable balanced flow matching baselines: Independent CFM (I-CFM) \citep{cfm_lipman}, which independently samples source and target cells, and OT-CFM \citep{cfm_tong}, which computes a balanced OT coupling within each minibatch of 256 cells during training. We evaluate MUST-FM both with and without biological transition priors under a leave-one-out setting, where one intermediate stage is held out and predicted from the remaining observed stages. 

As shown in Table~\ref{tab:q3_moca_prior}, OT-CFM consistently outperforms I-CFM, highlighting the importance of optimal transport for single-cell trajectory reconstruction. Without biological priors, MUST-FM achieves comparable performance to OT-CFM at E10.5 and E11.5 and improves the prediction at E12.5, resulting in a lower average 1-Wasserstein distance. This suggests that explicitly modeling unbalanced mass variation, including cell proliferation and death, can further improve reconstruction beyond balanced OT. Incorporating biological transition priors further reduces the 1-Wasserstein distance at E10.5 and E11.5, yielding the best average performance at held-out stages, indicating improved interpolation accuracy at unobserved developmental time points. To further assess the robustness of these interpolation results, we conduct a hyperparameter sensitivity analysis on MOCA, varying the batch size, learning rate, and number of training epochs, as shown in Appendix~\ref{app: SA}.

\textbf{Scalability on Atlas-Scale Data (Q4).}
We evaluate the scalability of MUST-FM on MOCA by subsampling the atlas to $1\%$, $5\%$, $10\%$, $20\%$, $50\%$, and $100\%$ of the original cells, and compare two finest-level coupling strategies: sparse exact OET and final-step within-block independent coupling. We report coupling construction time, flow matching training time, total runtime, peak CPU memory, and peak GPU memory. 

As shown in Figure~\ref{fig:q4_scalability}, sparse exact OET becomes infeasible at larger scales due to memory limitations, whereas independent lifting scales to the full MOCA atlas. Under independent lifting, coupling construction grows much more slowly than the overall training time, so its fraction of total runtime decreases as data size increases. Peak CPU and GPU memory also remain nearly stable across scales, because coupling construction is performed at annotation levels whose number of groups is much smaller than the number of cells. These results show that final-step independent lifting enables MUST-FM to scale to atlas-level single-cell datasets without explicitly constructing the finest-level coupling.

\section{Conclusion, Limitations, and Discussion}

We proposed MUST-FM, a multiscale supervised unbalanced optimal transport flow matching framework for learning population dynamics from single-cell and spatial omics data. By combining transition priors, masked optimal entropy transport, hierarchical coupling construction, and simulation-free flow matching, MUST-FM learns both transport and growth dynamics while avoiding dense point-level transport computation at the finest resolution. A detailed comparison with existing trajectory inference methods and a further discussion of the technical contributions of MUST-FM are provided in Appendix~\ref{app:F}.

Our experiments demonstrate that the multiscale strategy can recover hierarchical coupling structures with substantially reduced computational cost. The sparse variant achieves near-exact coupling recovery on synthetic benchmarks, while the independent lifting variant offers an efficient alternative when the goal is to learn coarse-to-mesoscopic dynamics rather than exact point-level matching.

MUST-FM still has several limitations. Its performance depends on the quality of the provided hierarchy and transition priors, and inaccurate priors may either remove valid transitions or introduce implausible ones. In addition, the independent lifting approximation sacrifices point-level coupling accuracy for scalability. Finally, the current formulation uses transition priors as predefined constraints; moving beyond hard constraints by learning, refining, or adaptively relaxing these priors from data is an important direction for future work.

\begin{ack}
P.Z. acknowledges support from the National Natural Science Foundation of China under grants 12288101, 8206100646, and T2321001, the Fundamental and Interdisciplinary Disciplines Breakthrough Plan of the Ministry of Education of China under grant JYB2025XDXM502, and the Fundamental Research Funds for the Central Universities. We also acknowledge the support from the High-performance Computing Platform of Peking University for computation.
\end{ack}

%
%
%
%
%
%
%
%
%
%

\bibliographystyle{unsrt} 
\bibliography{references} 

\newpage

\appendix

\section{Appendix A}
\subsection{Implicit Finest-Level Semi-Coupling Sampling and WFR Geodesic Targets}
\label{app:implicit_finest_sampling}

In this section, we describe how to perform flow-matching sampling at the finest level without explicitly materializing the finest-level coupling or its semi-couplings. We emphasize that, throughout this section, $\gamma_0$ denotes the \emph{starting semi-coupling} rather than the source marginal. Thus, $\gamma_0$ has the same shape as the coupling $\gamma$.

\paragraph{Penultimate-level supervised OET coupling and semi-couplings.}
Assume that the supervised OET problem is solved only up to the penultimate level $L-1$. Let
\[
\gamma^{(L-1)} \in \mathbb{R}_+^{K_0^{(L-1)} \times K_1^{(L-1)}}
\]
denote the resulting penultimate-level coupling. For notational simplicity, define the row and column sums
\[
s_I = \sum_{J} \gamma_{IJ}^{(L-1)},
\qquad
t_J = \sum_{I} \gamma_{IJ}^{(L-1)}.
\]
By the semi-coupling construction, the corresponding starting and ending semi-couplings at level $L-1$ are
\begin{equation}
\label{eq:coarse_semicouplings}
\gamma_{0,IJ}^{(L-1)}
=
\frac{\gamma_{IJ}^{(L-1)}}{s_I}
w_{0,I}^{(L-1)},
\qquad
\gamma_{1,IJ}^{(L-1)}
=
\frac{\gamma_{IJ}^{(L-1)}}{t_J}
w_{1,J}^{(L-1)}.
\end{equation}
Here $w_{0,I}^{(L-1)}$ and $w_{1,J}^{(L-1)}$ are the empirical masses of the source and target parent clusters. We use the convention that entries with zero denominators are ignored, since they carry no mass in the corresponding semi-coupling.

\paragraph{Implicit finest-level lifting.}
For each finest-level source element $i$ and target element $j$, let
\[
I=\operatorname{pa}_0(i),
\qquad
J=\operatorname{pa}_1(j)
\]
denote their parent clusters at level $L-1$. Let $\mathcal{C}_0(I)$ and $\mathcal{C}_1(J)$ denote the finest-level children of source parent $I$ and target parent $J$, respectively. The parent weights satisfy
\[
w_{0,I}^{(L-1)}
=
\sum_{i\in \mathcal{C}_0(I)} w_{0,i}^{(L)},
\qquad
w_{1,J}^{(L-1)}
=
\sum_{j\in \mathcal{C}_1(J)} w_{1,j}^{(L)}.
\]
Define the within-block empirical sampling probabilities
\begin{equation}
\label{eq:child_probs}
\alpha_{i\mid I}
=
\frac{w_{0,i}^{(L)}}{w_{0,I}^{(L-1)}},
\qquad
i\in \mathcal{C}_0(I),
\end{equation}
and
\begin{equation}
\label{eq:child_probs_target}
\beta_{j\mid J}
=
\frac{w_{1,j}^{(L)}}{w_{1,J}^{(L-1)}},
\qquad
j\in \mathcal{C}_1(J).
\end{equation}
These probabilities satisfy
\[
\sum_{i\in \mathcal{C}_0(I)} \alpha_{i\mid I}=1,
\qquad
\sum_{j\in \mathcal{C}_1(J)} \beta_{j\mid J}=1.
\]

Under the within-block independence assumption, the implicit finest-level lifted coupling is
\begin{equation}
\label{eq:implicit_lifted_gamma}
\widetilde{\gamma}_{ij}^{(L)}
=
\gamma_{IJ}^{(L-1)}
\alpha_{i\mid I}
\beta_{j\mid J},
\qquad
I=\operatorname{pa}_0(i),\quad J=\operatorname{pa}_1(j).
\end{equation}
Equivalently,
\[
\widetilde{\gamma}_{ij}^{(L)}
=
\gamma_{IJ}^{(L-1)}
\frac{w_{0,i}^{(L)}}{w_{0,I}^{(L-1)}}
\frac{w_{1,j}^{(L)}}{w_{1,J}^{(L-1)}}.
\]
In practice, we never construct $\widetilde{\gamma}^{(L)}$ explicitly. It is only used to define the implicit sampling distribution.

\paragraph{Induced finest-level semi-couplings.}
We next show that the finest-level starting semi-coupling induced by \eqref{eq:implicit_lifted_gamma} is
\begin{equation}
\label{eq:implicit_lifted_gamma0}
\widetilde{\gamma}_{0,ij}^{(L)}
=
\gamma_{0,IJ}^{(L-1)}
\alpha_{i\mid I}
\beta_{j\mid J},
\qquad
I=\operatorname{pa}_0(i),\quad J=\operatorname{pa}_1(j).
\end{equation}
Similarly, the induced ending semi-coupling is
\begin{equation}
\label{eq:implicit_lifted_gamma1}
\widetilde{\gamma}_{1,ij}^{(L)}
=
\gamma_{1,IJ}^{(L-1)}
\alpha_{i\mid I}
\beta_{j\mid J}.
\end{equation}

\paragraph{Derivation of the starting semi-coupling.}
By definition of the starting semi-coupling, we have
\begin{equation}
\label{eq:finest_gamma0_definition}
\widetilde{\gamma}_{0,ij}^{(L)}
=
\frac{
\widetilde{\gamma}_{ij}^{(L)}
}{
\sum_{j'} \widetilde{\gamma}_{ij'}^{(L)}
}
w_{0,i}^{(L)}.
\end{equation}
We first compute the denominator. For fixed $i$ with parent $I=\operatorname{pa}_0(i)$,
\begin{align}
\sum_{j'} \widetilde{\gamma}_{ij'}^{(L)}
&=
\sum_{J'}
\sum_{j'\in \mathcal{C}_1(J')}
\gamma_{I J'}^{(L-1)}
\alpha_{i\mid I}
\beta_{j'\mid J'}
\\
&=
\alpha_{i\mid I}
\sum_{J'}
\gamma_{I J'}^{(L-1)}
\sum_{j'\in \mathcal{C}_1(J')}
\beta_{j'\mid J'}
\\
&=
\alpha_{i\mid I}
\sum_{J'}
\gamma_{I J'}^{(L-1)}
\\
&=
\alpha_{i\mid I} s_I.
\end{align}
Substituting this and \eqref{eq:implicit_lifted_gamma} into \eqref{eq:finest_gamma0_definition} gives
\begin{align}
\widetilde{\gamma}_{0,ij}^{(L)}
&=
\frac{
\gamma_{IJ}^{(L-1)}
\alpha_{i\mid I}
\beta_{j\mid J}
}{
\alpha_{i\mid I}s_I
}
w_{0,i}^{(L)}
\\
&=
\frac{\gamma_{IJ}^{(L-1)}}{s_I}
\beta_{j\mid J}
w_{0,i}^{(L)}
\\
&=
\frac{\gamma_{IJ}^{(L-1)}}{s_I}
w_{0,I}^{(L-1)}
\frac{w_{0,i}^{(L)}}{w_{0,I}^{(L-1)}}
\beta_{j\mid J}
\\
&=
\gamma_{0,IJ}^{(L-1)}
\alpha_{i\mid I}
\beta_{j\mid J}.
\end{align}
This proves \eqref{eq:implicit_lifted_gamma0}. The proof of \eqref{eq:implicit_lifted_gamma1} is analogous, using the ending semi-coupling definition
\[
\widetilde{\gamma}_{1,ij}^{(L)}
=
\frac{
\widetilde{\gamma}_{ij}^{(L)}
}{
\sum_{i'} \widetilde{\gamma}_{i'j}^{(L)}
}
w_{1,j}^{(L)}.
\]

\paragraph{Sampling from the finest-level starting semi-coupling.}
The unbalanced flow-matching objective samples pairs from the starting semi-coupling,
\[
(\bm{x}_0,\bm{x}_1)\sim \widetilde{\gamma}_{0}^{(L)}.
\]
Equation \eqref{eq:implicit_lifted_gamma0} shows that this sampling can be performed without explicitly storing $\widetilde{\gamma}_{0}^{(L)}$. Specifically, we use the following hierarchical procedure:
\begin{equation}
\label{eq:implicit_sampling_procedure}
\boxed{
\begin{aligned}
&(I,J) \sim \gamma_{0}^{(L-1)},\\
&i \sim \alpha_{\cdot\mid I}
=
\frac{w_{0,\cdot}^{(L)}}{w_{0,I}^{(L-1)}}
\quad \text{within } \mathcal{C}_0(I),\\
&j \sim \beta_{\cdot\mid J}
=
\frac{w_{1,\cdot}^{(L)}}{w_{1,J}^{(L-1)}}
\quad \text{within } \mathcal{C}_1(J),\\
&\bm{x}_0=\bm{x}_{0,i}^{(L)},
\qquad
\bm{x}_1=\bm{x}_{1,j}^{(L)}.
\end{aligned}
}
\end{equation}
Indeed, the probability of sampling a finest-level pair $(i,j)$ is
\begin{align}
\mathbb{P}(i,j)
&=
\mathbb{P}(I,J)
\mathbb{P}(i\mid I)
\mathbb{P}(j\mid J)
\\
&\propto
\gamma_{0,IJ}^{(L-1)}
\alpha_{i\mid I}
\beta_{j\mid J}
\\
&=
\widetilde{\gamma}_{0,ij}^{(L)}.
\end{align}
Thus, the procedure in \eqref{eq:implicit_sampling_procedure} is exactly equivalent to sampling from the finest-level starting semi-coupling. If the semi-coupling is not normalized to have total mass one, the implementation samples from the normalized distribution proportional to $\gamma_0^{(L-1)}$. The missing total mass is a constant factor in the Monte Carlo estimator and can either be multiplied back or ignored when it does not affect the optimizer.

\paragraph{Relative endpoint masses for the WFR traveling Dirac.}
Since the flow-matching objective samples from the starting semi-coupling $\widetilde{\gamma}_{0}^{(L)}$, we use a relative mass parametrization for each sampled path. For a sampled finest-level pair $(i,j)$ with parent pair $(I,J)$, set
\begin{equation}
\label{eq:relative_endpoint_masses}
m_0 = 1,
\qquad
m_1 =
\rho_{IJ}
:=
\frac{
\widetilde{\gamma}_{1,ij}^{(L)}
}{
\widetilde{\gamma}_{0,ij}^{(L)}
}.
\end{equation}
Using \eqref{eq:implicit_lifted_gamma0} and \eqref{eq:implicit_lifted_gamma1}, the child weights cancel:
\begin{align}
\rho_{IJ}
&=
\frac{
\gamma_{1,IJ}^{(L-1)}
\alpha_{i\mid I}
\beta_{j\mid J}
}{
\gamma_{0,IJ}^{(L-1)}
\alpha_{i\mid I}
\beta_{j\mid J}
}
\\
&=
\frac{\gamma_{1,IJ}^{(L-1)}}{\gamma_{0,IJ}^{(L-1)}}.
\end{align}

\paragraph{Monte Carlo estimator for flow matching.}
For each mini-batch sample, we first draw
\[
(I,J)\sim \gamma_0^{(L-1)},\qquad
i\sim \alpha_{\cdot\mid I},\qquad
j\sim \beta_{\cdot\mid J}.
\]
We then set
\[
\bm{x}_0=\bm{x}_{0,i}^{(L)},
\qquad
\bm{x}_1=\bm{x}_{1,j}^{(L)},
\qquad
m_0=1,
\qquad
m_1=\rho_{IJ}.
\]
Next, sample
\[
t\sim \mathcal{U}[0,1],
\qquad
\bm{x}\sim \mathcal{N}(\bm{\eta}_t,\sigma_t^2\mathbf{I}).
\]

The WFR geodesic targets $(\bm{u}_t,g_t,m(t))$ are then computed using the traveling-Dirac formulas introduced in the main text. The corresponding mini-batch objective is
\begin{equation}
\label{eq:implicit_cufm_objective}
\mathcal{L}_{\mathrm{CUFM}}
=
\mathbb{E}
\left[
\left(
\|\bm{v}_{\bm{\theta}}(\bm{x},t)-\bm{u}_t\|_2^2
+
\kappa
\|g_{\bm{\phi}}(\bm{x},t)-g_t\|_2^2
\right)
m(t)
\right],
\end{equation}
where the expectation is taken over the hierarchical sampling procedure above. Importantly, this estimator is equivalent to sampling
\[
(\bm{x}_0,\bm{x}_1)\sim \widetilde{\gamma}_0^{(L)}
\]
from the implicit finest-level starting semi-coupling, but it only requires storing the penultimate-level semi-coupling $\gamma_0^{(L-1)}$, the parent-child assignments, and the within-parent empirical weights. Therefore, neither the finest-level coupling $\widetilde{\gamma}^{(L)}$ nor its semi-couplings $\widetilde{\gamma}_0^{(L)}$ and $\widetilde{\gamma}_1^{(L)}$ need to be explicitly constructed or stored, which avoids the memory cost of enumerating all finest-level source-target pairs.

\section{Appendix B}
\subsection{Pseudocode of MUST-FM}
\label{subsec:pseudocode}
We outline the unified training procedure for multi-time point trajectory inference in Algorithm \ref{alg:MUST-FM}.

\begin{algorithm}[h]
	\caption{MUST-FM Training Workflow}
	\label{alg:MUST-FM}
	\begin{algorithmic}[1]
        \Require Time-series datasets $\{\mu_{t_k}\}_{k=0}^K$, $L$-level hierarchical annotations, biological priors $\{B^{(l)}\}_{l=1}^{L-1}$, transition threshold $\epsilon$, Gaussian bandwidth $\sigma$, batch size $b$, networks $\bm{v}_{\bm{\theta}}$ and $g_{\bm{\phi}}$.
		\State \textbf{/* Phase 1: Full-Batch Multiscale Pre-computation */}
		\For{$k = 0 \to K-1$}
		\State Extract multiscale centroids and weights for $(\mu_{t_k}, \mu_{t_{k+1}})$ using hierarchical annotations.
		\State Compute supervised OET couplings for levels $1$ to $L-1$. \hfill \Comment{Eq. \ref{eq:final_mask} and Eq. \ref{eq:discrete_OET}}
		
		\State \textbf{/* Final-step handling of the finest level */}
		\State \textbf{Option 1 (Exact Sparse OET at the finest level):}
		\State \quad $\gamma^{(k)} \gets \text{Sparse OET}(\hat{\mu}_{t_k}^{(L)}, \hat{\mu}_{t_{k+1}}^{(L)} \mid M^{(k,L)})$ \hfill \Comment{Eq. \ref{eq:discrete_OET}}
		\State \quad Construct semi-coupling $\gamma_0^{(k)}, \gamma_1^{(k)}$ from the finest-level coupling $\gamma^{(k)}$. \hfill \Comment{Theorem \ref{thm:semi-coupling}}
		
		\State \textbf{Option 2 (Final-Step Within-Block Independent Coupling Lifting):}
		\State \quad Store $\gamma^{(k,L-1)}$, finest-level weights, and parent mappings for implicit finest-level sampling. \hfill \Comment{Eq. \ref{eq:coarse_to_fine_lifting}; Appendix \ref{app:implicit_finest_sampling}}
		\EndFor
		
		\State \textbf{/* Phase 2: Simulation-Free Continuous Flow Training */}
		\While{Training not converged}
		\For{$k = 0 \to K-1$}
		\State \textbf{If Option 1 is used:}
		\State \quad Sample paired cells $(\bm{x}_{t_k}, \bm{x}_{t_{k+1}}) \sim \gamma_0^{(k)}$ with batch size $b$.
		\State \quad Set $\bar{\gamma}_0^{(k)} \gets \gamma_0^{(k)}(\bm{x}_{t_k},\bm{x}_{t_{k+1}})$ and $\bar{\gamma}_1^{(k)} \gets \gamma_1^{(k)}(\bm{x}_{t_k},\bm{x}_{t_{k+1}})$.
		\State \textbf{If Option 2 is used:}
		\State \quad Sample $(\bm{x}_{t_k},\bm{x}_{t_{k+1}})$ and on-the-fly semi-coupling values $\bar{\gamma}_0^{(k)},\bar{\gamma}_1^{(k)}$ from the implicit lifted sampler. \hfill \Comment{No explicit level-$L$ semi-coupling matrix. Appendix \ref{app:implicit_finest_sampling}}
		\State Set mass boundaries: $m_0 = 1$ and $m_1 = \bar{\gamma}_1^{(k)} / \bar{\gamma}_0^{(k)}$.
		\State Compute geometric constants $A, B, \bm{\omega}_0$ and $\tau$ for the WFR geodesic.
		\State Sample normalized time $t \sim \mathcal{U}(0,1)$, set absolute time $t^{(k)} \gets t_k + (t_{k+1}-t_k)t$.
		
		\State \textbf{/* Calculate conditional path targets */}
		\State $\bm{\eta}_{t^{(k)}} \gets \bm{x}_{t_k} +\bm{\omega}_0 \Lambda_t(\bm{x}_{t_k},\bm{x}_{t_{k+1}})$ \hfill \Comment{Eq. \ref{eq:travelling_dirac} and Eq. \ref{eq:travelling_gaussian_mean}}
		\State $\bm{x}^{(k)} \sim \mathcal{N}(\bm{\eta}_{t^{(k)}}, \sigma^2 \mathbf{I})$ \hfill \Comment{Eq. \ref{eq:CGMP}}
		\State $\bm{u}^{(k)} \gets \bm{\omega}_0/(m_t(\bm{x}_{t_k},\bm{x}_{t_{k+1}})(t_{k+1}-t_k))$ \hfill \Comment{Eq. \ref{eq:travelling_dirac}}
		\State $g^{(k)} \gets \frac{\mathrm{d}}{\mathrm{d}t} \ln m_t(\bm{x}_{t_k},\bm{x}_{t_{k+1}})/(t_{k+1}-t_k)$ \hfill \Comment{Eq. \ref{eq:travelling_dirac}}
		\State $m^{(k)} \gets m_t(\bm{x}_{t_k},\bm{x}_{t_{k+1}})/\bar{\gamma}_0^{(k)}$ \hfill \Comment{Eq. \ref{eq:travelling_dirac} and Eq. \ref{eq:travelling_gaussian_mean}}
		\EndFor
		
		\State Concatenate $\{\bm{x}^{(i)}, t^{(i)}, \bm{u}^{(i)}, g^{(i)}, m^{(i)}\}_{i=1}^K$ into batch tensors $\{\bm{x}^{\text{c}}, t^{\text{c}}, \bm{u}^{\text{c}}, g^{\text{c}}, m^{\text{c}}\}$.
		\State Calculate mass-weighted regression loss:
		\State $\mathcal{L}_{\text{CUFM}} \gets \frac{1}{b\cdot K}\sum \Big( \left\| \bm{v}_{\bm{\theta}}(\bm{x}^{\text{c}}, t^{\text{c}}) - \bm{u}^{\text{c}} \right\|_2^2 + \kappa \left\| g_{\bm{\phi}}(\bm{x}^{\text{c}}, t^{\text{c}}) - g^{\text{c}}\right\|_2^2 \Big) m^{\text{c}}$  \hfill \Comment{Eq. \ref{eq:CUFM_simplified}}
		\State $\bm{\theta}, \bm{\phi} \gets \text{AdamOptimizer}(\nabla_{\bm{\theta}} \mathcal{L}_{\text{CUFM}}, \nabla_{\bm{\phi}} \mathcal{L}_{\text{CUFM}})$
		\EndWhile
		\State \Return $\bm{v}_{\bm{\theta}}$ and $g_{\bm{\phi}}$
	\end{algorithmic}
\end{algorithm}

\section{Appendix C}
\label{app:E}

\subsection{Information on the computational resources}
All the experiments were performed on a RTX PRO 6000 (96GB) GPU. 

\subsection{Selection of hyperparameters}

\begin{table}[ht]
\centering
\caption{\textbf{Selection of hyperparameters across experiments.}}
\label{tab:hyperparameters}
\begin{tabular}{lccccccc}
\toprule
Experiment & Dim & $\delta$ & Batch size & Learning rate & Independent & Prior & Epochs \\
\midrule
EB 50D & 50 & 25.0 & 512 & 0.005 & \xmark & \xmark & 3000  \\
Simulation 2D & 2 & 1.5 & 400 & 0.005 & \xmark & \xmark & 3000 \\
Gaussian 1000D & 1000 & 1.4 & 256 & 0.001 & \xmark & \xmark & 3000 \\
Weinreb 2D & 2 & 15.0 & 512 & 0.005 & \xmark & \xmark & 3000 \\
MOCA 10D & 10 & 25.0 & 256 & 0.005 & \cmark & \cmark & 3000 \\
\bottomrule
\end{tabular}
\end{table}

The velocity field $\bm{v}_{\bm{\theta}}(\bm{x},t)$ and the growth-rate function $g_{\bm{\phi}}(\bm{x},t)$ were both parameterized by multilayer perceptrons with 5 layers and 256 hidden channels, using LeakyReLU activation functions. The selection of hyperparameters is shown in Table~\ref{tab:hyperparameters}. For the WFR geometry, the parameter $\delta$ controls the relative scale between spatial transport and mass creation or annihilation. For a fair comparison, we set $\delta$ on each dataset to be consistent with the setting used by WFR-FM~\citep{peng2026wfrfm}. Therefore, $\delta$ was not additionally tuned for our MUST-FM, the performance differences mainly reflect the effect of multiscale coupling construction and transition-prior supervision rather than dataset-specific adjustment of the WFR geometry.

\section{Appendix D}

\subsection{Multiscale Synthetic Dataset}
\label{app:balanced_synthetic_data}

We constructed a two-dimensional synthetic dataset with an explicit hierarchical cluster structure to examine whether leveraging intrinsic multiscale organization improves transport accuracy when such structure is present in the data. The dataset consists of two time points, denoted by $\mu_0$ and $\mu_1$, where $\mu_1$ is generated by translating $\mu_0$ along the horizontal direction. Therefore, the total mass and the cluster-wise mass are exactly preserved between the two time points.

The initial distribution $\mu_0$ contains two annotation levels: a coarse macro-cluster level and a fine micro-cluster level. Specifically, we generate three macro clusters, indexed by $m\in\{0,1,2\}$, corresponding to the top, middle, and bottom groups. Their centers are placed at
\begin{equation}
    \bm{q}_m = (a, b + d_m),
    \quad
    d_m \in \{5,0,-5\},
\end{equation}
where we set $a=2$ and $b=2$ in the experiments. Each macro cluster contains nine micro clusters arranged in a $3\times 3$ local pattern. The relative offsets of the nine micro-cluster centers are
\begin{equation}
    \mathcal{O}
    =
    \{(0,0),(0,1),(0,-1),(-1,0),(1,0),(1,1),(1,-1),(-1,1),(-1,-1)\}.
\end{equation}
Thus, for macro cluster $m$ and local micro-cluster index $r$, the corresponding micro-cluster center is
\begin{equation}
    \bm{c}_{m,r}^{(0)} = \bm{q}_m + \bm{o}_r,
    \quad
    \bm{o}_r \in \mathcal{O}.
\end{equation}

For each micro cluster, we sample $N$ points from an isotropic Gaussian distribution:
\begin{equation}
    \bm{x}_{m,r,n}^{(0)}
    \sim
    \mathcal{N}\left(\bm{c}_{m,r}^{(0)}, \sigma^2 \mathbf{I}_2\right),
    \quad
    n=1,\ldots,N.
\end{equation}
In our experiments, we use $N=1000$ and $\sigma=0.1$. This gives $3$ macro clusters, $27$ micro clusters, and $27{,}000$ cells at the first time point.

\begin{figure}[t]
    \centering
    \includegraphics[width=0.95\linewidth]{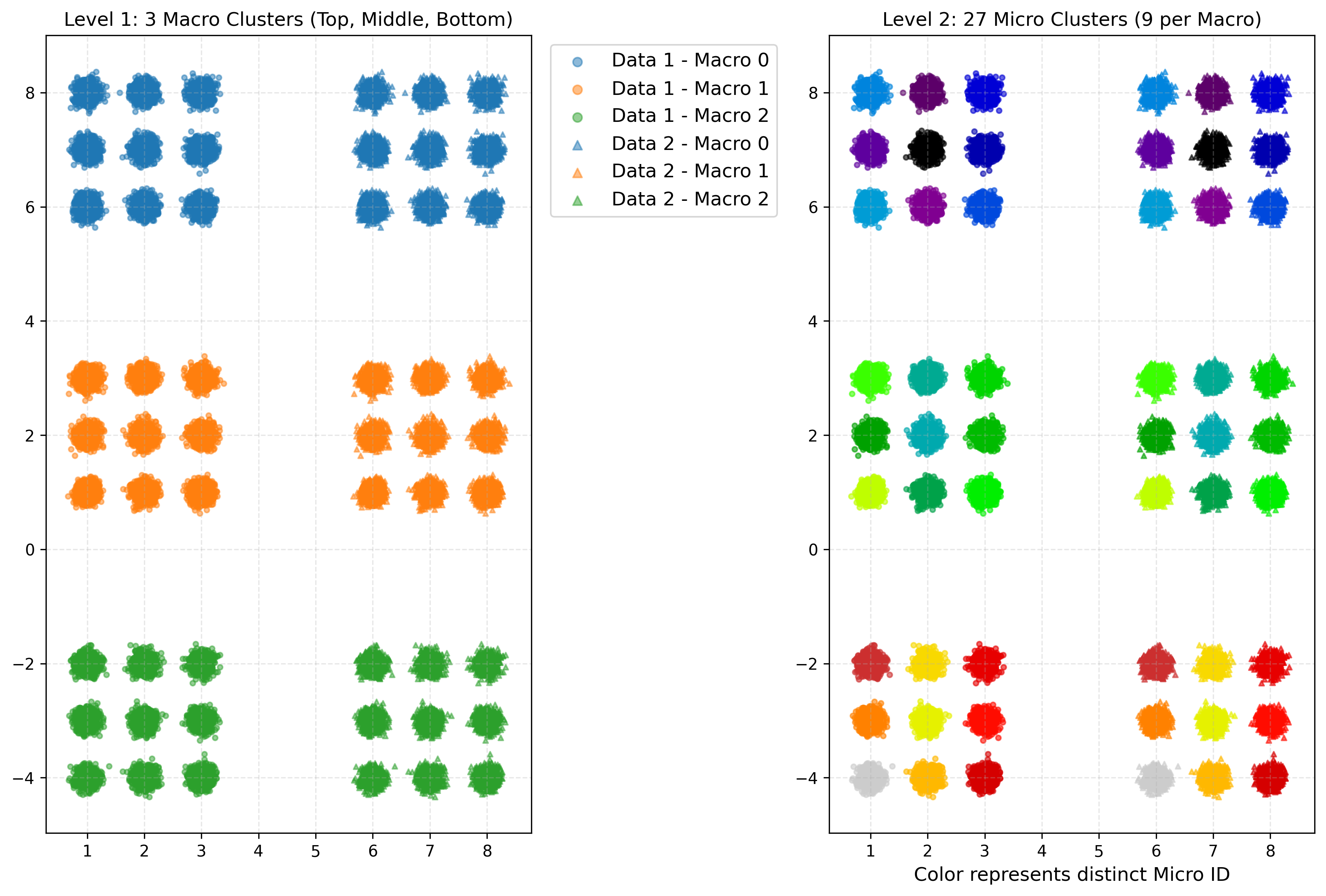}
    \caption{
    \textbf{Multiscale balanced synthetic dataset.}
    The dataset contains two time points with identical hierarchical cluster structure. 
    The left panel visualizes the coarse macro-cluster level, where three macro clusters are arranged vertically and the second time point is obtained by translating the first time point to the right.
    The right panel visualizes the fine micro-cluster level, where each macro cluster contains nine micro clusters arranged in a local $3\times 3$ pattern.
    Circles and triangles denote the two time points, respectively.
    }
    \label{fig:balanced_synthetic_data}
\end{figure}

The second time point is generated by applying a global translation to all cells:
\begin{equation}
    \bm{x}_{m,r,n}^{(1)}
    =
    \bm{x}_{m,r,n}^{(0)}
    +
    (5,0).
\end{equation}
The macro-cluster and micro-cluster labels are kept unchanged after the translation. Therefore, the ground-truth transport map is
\begin{equation}
    T(\bm{x}) = \bm{x} + (5,0),
\end{equation}
and the ground-truth velocity is a constant horizontal displacement field. Since every source cell has a corresponding translated target cell and all cluster sizes are preserved, this simulation represents a balanced multiscale transport setting without birth or death dynamics. An intuitive presentation of the simulated data is shown in Figure~\ref{fig:balanced_synthetic_data}.

The final dataset contains both time points and stores, for each cell, its two-dimensional coordinates, macro-cluster label, micro-cluster label, and time-point indicator. This construction provides a controlled benchmark in which the correct transport structure is known at both the global and hierarchical annotation levels.

\subsection{Metrics for Controlled Synthetic Experiment}
\label{app:metric}

Because the synthetic dataset provides a known ground-truth transport correspondence at three hierarchical resolutions: the point, micro-cluster, and macro-cluster levels, we adopt specific metrics for our controlled synthetic experiment. Let $\gamma^{\mathrm{GT}}$ denote the ground-truth coupling,
$\widehat{\gamma}$ the coupling estimated by a method, and
$\mathcal{J}(\gamma)$ the transport objective evaluated using the same
finest-level cost and regularization. The relative objective gap is
defined as
\begin{equation}
\operatorname{ObjGap}(\widehat{\gamma})
=
\frac{
\left|\mathcal{J}(\widehat{\gamma})
-\mathcal{J}(\gamma^{\mathrm{GT}})\right|
}{
\left|\mathcal{J}(\gamma^{\mathrm{GT}})\right|
}
\times 100\%.
\end{equation}

For the hierarchical accuracy metrics, let $T(i)$ denote the
ground-truth target corresponding to source point $i$, and let
$c_{\mathrm{micro}}(j)$ and $c_{\mathrm{macro}}(j)$ denote the
micro- and macro-cluster labels of target point $j$, respectively.
We measure the fraction of the estimated transport mass assigned
to a ground-truth-consistent target at each resolution:

\begin{equation}
\operatorname{PointAcc}(\widehat{\gamma})
=
\frac{
\sum_i \widehat{\gamma}_{i,T(i)}
}{
\sum_{i,j}\widehat{\gamma}_{ij}
},
\end{equation}

\begin{equation}
\operatorname{MicroAcc}(\widehat{\gamma})
=
\frac{
\sum_{i,j}\widehat{\gamma}_{ij}
\mathbf{1}\left[
c_{\mathrm{micro}}(j)
=
c_{\mathrm{micro}}(T(i))
\right]
}{
\sum_{i,j}\widehat{\gamma}_{ij}
},
\end{equation}

and

\begin{equation}
\operatorname{MacroAcc}(\widehat{\gamma})
=
\frac{
\sum_{i,j}\widehat{\gamma}_{ij}
\mathbf{1}\left[
c_{\mathrm{macro}}(j)
=
c_{\mathrm{macro}}(T(i))
\right]
}{
\sum_{i,j}\widehat{\gamma}_{ij}
}.
\end{equation}

\section{Appendix E}
\label{app:D}

\subsection{Theoretical analysis of the multiscale approximation}
\label{app:D1}

We conduct theoretical analysis to clarify when the multiscale approximation is reliable. The central intuition is that hierarchical pruning is less likely to discard correct transitions when the data are well aligned with the assumed multiscale structure. By ``well aligned,'' we mean that points within each annotated cluster are concentrated around their centroid, while different clusters corresponding to competing transitions are well separated relative to their within-cluster variation. In this regime, cluster centroids accurately represent their constituent cells, and the coarse coupling provides reliable guidance for the finer levels.

This intuition can be formalized through a sufficient condition for preserving important transitions. Let $q_{IJ}^\star$ denote the proportion of source-cluster mass transported from parent cluster $I$ to $J$, obtained by aggregating the full fine-scale solution. Let $\widehat q_{IJ}$ be the corresponding proportion estimated by the coarse solution, and define the maximum coarse approximation error as

\begin{equation}
\kappa = \max_{I,J} \left| \widehat q_{IJ}-q_{IJ}^\star \right|.
\end{equation}

Let $p_{\min}$ be the smallest transport proportion among the meaningful transitions that should be preserved. If the biological mask does not forbid these transitions and

\begin{equation}
\epsilon+\kappa<p_{\min},
\end{equation}

then none of them can be removed by the hierarchical mask. Indeed, for every meaningful transition,

\begin{equation}
\widehat q_{IJ}
\ge
q_{IJ}^\star-\kappa
\ge
p_{\min}-\kappa >
\epsilon,
\end{equation}

so its child-level transport block remains admissible. If this condition holds recursively at every hierarchy level, all important transport paths are preserved throughout the coarse-to-fine procedure. If these paths contain the complete support of the fine-scale optimum, the masked and unmasked problems have the same optimal value; if the optimum is unique, they also have the same coupling.

The quantity $\kappa$ is controlled by how well the data satisfy the multiscale structure described above. Let $r_0$ and $r_1$ be the maximum source- and target-cluster radii, and assume that the transport cost is $L_c$-Lipschitz over the relevant distance range. For a fixed set of admissible parent blocks, replacing individual points with cluster centroids introduces an objective error controlled by

\begin{equation}
L_c(r_0+r_1)(m_f^\star+m_c^\star),
\end{equation}

where $m_f^\star$ and $m_c^\star$ are the transported masses of the fine and coarse solutions. If the coarse problem additionally satisfies a local stability condition with constant $\lambda>0$, then

\begin{equation}
\kappa
\le
\frac{
L_c(r_0+r_1)(m_f^\star+m_c^\star)
}{
\lambda w_{\min}
},
\end{equation}

where $w_{\min}$ is the smallest source-cluster mass. Here, compact clusters lead to smaller $r_0+r_1$, while better separation between competing cluster-level transitions leads to a more stable coarse solution and therefore a larger $\lambda$. Both effects reduce $\kappa$, making the support-preservation condition easier to satisfy.

The same cluster compactness also controls the additional approximation introduced by MUST-FM (Independent). If $\mathcal L\Gamma_c^\star$ denotes the within-block independent lifting in Eq.~15, then

\begin{equation}
0
\le
F_f(\mathcal L\Gamma_c^\star)-F_f^\star
\le
L_c(r_0+r_1)(m_f^\star+m_c^\star).
\end{equation}

Thus, when the data exhibit a clear multiscale structure, both hierarchical pruning and within-block lifting introduce limited error. Conversely, diffuse clusters, poorly separated competing transitions, rare true transitions, an overly aggressive threshold, or an inaccurate biological mask may cause important fine-scale structure to be lost.

\subsection{Empirical study of the multiscale approximation}
\label{app:D2}

We further provide empirical validation through two controlled stress tests in the three-level synthetic setting, consisting of point, micro, and macro levels. The full unmasked fine-scale solution is used as the reference. Obj.Gap measures its relative objective difference from the hierarchical solution, while PointAcc., MicroAcc., and MacroAcc. measure the proportions of reference transitions recovered at the corresponding scales. The pruning threshold and all other parameters are held fixed.

The central intuition is that hierarchical pruning is reliable when the data exhibit a clear multiscale structure: points are concentrated within clusters, while competing clusters are well separated. The following experiments separately weaken these two properties.

\textbf{Increasing within-cluster variance:} Source and target cells are independently sampled from Gaussian distributions around their respective cluster centroids. Only the centroids follow the prescribed ground-truth transitions; target cells are not translated copies of source cells. We keep the centroids fixed and progressively increase the within-cluster variance. As shown in Table~\ref{tab:variance_sensitivity}, When the clusters are compact, the hierarchical solution exactly recovers the reference solution. As the variance increases, the cluster radii in our theoretical bound increase, making the coarse coupling less representative of fine-scale transitions. Accuracy consequently decreases first at the point level, then at the micro level, while the well-separated macro structure remains almost perfectly preserved.

\begin{table}[ht]
\centering
\caption{\textbf{Effect of increasing within-cluster variance.}}
\label{tab:variance_sensitivity}
\begin{tabular}{ccccc}
\toprule
Within-cluster variance & Obj.Gap(\%)$\downarrow$ & PointAcc.$\uparrow$ & MicroAcc.$\uparrow$ & MacroAcc.$\uparrow$ \\
\midrule
0.25 & 0.000 & 1.000 & 1.000 & 1.000 \\
0.5  & 0.000 & 1.000 & 1.000 & 1.000 \\
1.0  & 3.24e-4 & 0.483 & 0.825 & 1.000 \\
1.5  & 1.24e-3 & 0.241 & 0.556 & 1.000 \\
2.0  & 2.22e-3 & 0.175 & 0.422 & 0.997 \\
\bottomrule
\end{tabular}
\end{table}

\textbf{Decreasing micro-cluster separation:} We next keep the within-cluster variation and macro-level structure fixed but progressively reduce the separation between micro-cluster centroids corresponding to competing transitions. As shown in Table~\ref{tab:separation_sensitivity}, well-separated micro clusters are recovered exactly. Bringing competing centroids closer weakens the local stability term in our theoretical analysis, increasing the possible coarse-coupling error. Point- and micro-level accuracy therefore deteriorate, whereas MacroAcc. remains 1.000 because the macro structure is unchanged and well separated.

\begin{table}[ht]
\centering
\caption{\textbf{Effect of decreasing micro-cluster separation.}}
\label{tab:separation_sensitivity}
\begin{tabular}{ccccc}
\toprule
Micro-centroid separation & Obj.Gap(\%)$\downarrow$ & PointAcc.$\uparrow$ & MicroAcc.$\uparrow$ & MacroAcc.$\uparrow$ \\
\midrule
2.5 & 0.000 & 1.000 & 1.000 & 1.000 \\
2.0 & 0.000 & 1.000 & 1.000 & 1.000 \\
1.5 & 1.43e-7 & 0.993 & 0.999 & 1.000 \\
1.0 & 2.35e-5 & 0.671 & 0.948 & 1.000 \\
0.5 & 9.69e-5 & 0.290 & 0.587 & 1.000 \\
\bottomrule
\end{tabular}
\end{table}

These experiments directly support the theoretical prediction: hierarchical pruning preserves important transitions when within-cluster variation is small relative to between-cluster separation, but fine-scale transitions may be discarded as this multiscale structure becomes ambiguous. The objective gaps remain extremely small even when point-level accuracy decreases because, in the overlapping regime, many different fine-scale couplings have nearly identical costs. Thus, exact cell-to-cell assignments may become ambiguous while the coarse population-level transport remains accurate.

\section{Appendix F}
\subsection{Sensitivity analysis}
\label{app: SA}

We investigate the sensitivity of MUST-FM to three key training
hyperparameters: batch size, learning rate, and number of training
epochs, as shown in Table~\ref{tab:hyperparameter_sensitivity}. The reported MUST-FM model with biological priors uses a batch size of 256, a learning rate of 0.005, and 3,000 training epochs. We varied each hyperparameter independently while holding all other settings fixed. The reported $W_1$ is averaged over the three held-out stages E10.5, E11.5, and E12.5. Overall, MUST-FM is reasonably robust to these training choices. Batch sizes 256 and 512 perform comparably, whereas 128 is moderately worse; the three learning rates differ by at most 0.040 in average $W_1$, with 0.005 performing best. Similarly, 3,000 and 4,500 epochs yield nearly identical results, while 1,500 epochs is slightly less accurate. We therefore use 256, 0.005, and 3,000 epochs as a practical balance between accuracy and computational cost.

\begin{table}[h]
\centering
\caption{\textbf{Sensitivity analysis on MOCA dataset.}}
\label{tab:hyperparameter_sensitivity}
\small
\begin{tabular}{lcc}
\toprule
Hyperparameter & Selected value & Avg. $W_1$ $\downarrow$ \\
\midrule

\multirow{3}{*}{Batch size}
& 128 & 1.432 \\
& 256 & 1.336 \\
& 512 & 1.323 \\

\midrule

\multirow{3}{*}{Learning rate}
& 0.001 & 1.360 \\
& 0.005 & 1.336 \\
& 0.010 & 1.376 \\

\midrule

\multirow{3}{*}{Training epochs}
& 1,500 & 1.386 \\
& 3,000 & 1.336 \\
& 4,500 & 1.338 \\

\bottomrule
\end{tabular}
\end{table}

\section{Appendix G}
\label{app:F}

\subsection{Comparison with existing methods}

\begin{table}[ht]
\centering
\caption{\textbf{Comparison of existing trajectory inference algorithms.}}
\label{tab:comparison}
\begin{tabular}{lccccc}
\toprule
Algorithm & Unbalanced & Stochastic & Simulation-free & Prior  & Scalability\\
\midrule
TIGON \citep{TIGON} & \cmark & \xmark & \xmark & \xmark  & \xmark\\
DeepRUOT \citep{DeepRUOT} & \cmark & \cmark & \xmark & \xmark  & \xmark\\
WFR-FM \citep{peng2026wfrfm} & \cmark & \xmark & \cmark & \xmark  & \xmark\\
\midrule
MUST-FM & \cmark & \xmark & \cmark & \cmark  & \cmark\\
\bottomrule
\end{tabular}
\end{table}

Table~\ref{tab:comparison} summarizes key modeling and computational properties of representative trajectory inference methods. "Unbalanced" indicates whether the method explicitly models changes in total population mass, such as proliferation and apoptosis. "Stochastic" denotes whether the inferred dynamics are formulated as stochastic processes rather than deterministic flows. "Simulation-free" refers to training objectives that avoid repeated ODE simulation, typically through flow matching. "Prior" indicates whether biological transition constraints, lineage information, or other user-specified prior knowledge can be incorporated. "Scalability" measures whether the method is designed to scale to atlas-scale single-cell datasets.

The technical contributions of MUST-FM can be organized into three components that, to the best of our knowledge, have not been developed in prior work.

\textbf{1. Supervised unbalanced optimal transport.}
We introduce a supervised formulation of unbalanced optimal transport that incorporates prior transition constraints into both the dynamic WFR problem and its equivalent static optimal entropy-transport formulation. Although related to supervised balanced OT, the unbalanced setting introduces an important distinction in feasibility. In supervised balanced OT, a transition mask that is incompatible with the prescribed source and target marginals results in an empty feasible set. In our supervised unbalanced formulation, marginal discrepancies are penalized rather than prohibited. As a result, such a mask does not render the problem infeasible: unmatched source mass can be annihilated, while unmatched target mass can be created. This allows supervised UOT to remain mathematically well-defined under incompatible transition constraints and distinguishes it from a merely notational extension of supervised balanced OT.

\textbf{2. Multiscale unbalanced optimal transport.}
We introduce a coarse-to-fine multiscale construction specifically designed for UOT, which, to the best of our knowledge, is the first of its kind. Existing multiscale OT approaches have primarily focused on balanced transport. Our formulation constructs weighted measures at multiple annotation levels, combines biological transition masks with coupling-dependent coarse-to-fine masks, and solves supervised OET successively across these levels. In this way, transport and mass-variation information are propagated through the hierarchy, while the admissible search spaces are progressively restricted at finer resolutions.

\textbf{3. Implicit finest-level coupling and sampling.}
We develop a lifting and sampling strategy that enables cell-level training pairs for flow matching to be sampled without explicitly solving for or materializing the corresponding cell-level coupling. Specifically, the OET problem is solved only up to the penultimate annotation level, after which the resulting transport blocks are lifted implicitly according to the empirical within-block weights.



\end{document}